%% file: full.tex
\documentclass[a4paper,fleqn]{cas-dc}

\usepackage[sort,numbers]{natbib}
\usepackage[utf8]{inputenc} 
\usepackage[T1]{fontenc}    
\usepackage{hyperref}       
\usepackage{booktabs}       
\usepackage{amsfonts}       
\usepackage{nicefrac}       
\usepackage[letterspace=150]{microtype}
\usepackage{xcolor}         
\usepackage{graphicx}
\usepackage{subcaption}
\usepackage{amsmath}
\usepackage{amsmath,amssymb}
\usepackage{algorithmic}
\usepackage{graphicx}
\usepackage{textcomp}
\usepackage{datetime}
\usepackage{pythonhighlight}
\usepackage{algorithm} 
\usepackage{appendix}
\usepackage{mathtools}
\usepackage{wrapfig,booktabs}

\BeforeBeginEnvironment{appendices}{\clearpage}

\newcommand{\Defrag}{\emph{Defrag}\xspace}

\DeclareMathOperator*{\argmax}{argmax}

\DeclareMathOperator*{\std}{\sigma}

\providecommand{\tightlist}{%
  \setlength{\itemsep}{0pt}\setlength{\parskip}{0pt}}

\def\tsc#1{\csdef{#1}{\textsc{\lowercase{#1}}\xspace}}
\tsc{WGM}
\tsc{QE}

\begin{document}
\let\WriteBookmarks\relax
\def\floatpagepagefraction{1}
\def\textpagefraction{.001}

\shorttitle{Inferring Actual Treatment Pathways from Patient Records}

\shortauthors{Wilkins-Caruana et~al.}

\title [mode = title]{Inferring Actual Treatment Pathways from Patient Records}

%

%
\credit{Conceptualization of this study, Methodology, Software}
\author[1]{Adrian Wilkins-Caruana}[orcid=0000-0001-7283-0220]
    \cormark[1]
    \ead{adrian.caruana@uts.edu.au}
\author[1]{Madhushi Bandara}
\author[2]{Katarzyna Musial}
\author[1,3]{Daniel Catchpoole}
\author[1,4]{Paul J. Kennedy}






\affiliation[1]{
    organization={Australian Artificial Intelligence Institute, Faculty of Engineering and IT, University of Technology Sydney},
    country={Australia}
}
\affiliation[2]{
    organization={Complex Adaptive Systems Lab, Data Science Institute, Faculty of Engineering and IT, University of Technology Sydney},
	country={Australia}
}
\affiliation[3]{
    organization={Biospecimen Research Services, The Children's Cancer Research Unit},
    addressline={The Children's Hospital at Westmead}, 
	country={Australia}
}
\affiliation[4]{
    organization={Joint Research Centre in AI for Health and Wellness},
	country={University of Technology Sydney, Australia and Ontario Tech University, Canada}
}

\begin{abstract}%
    \emph{Objective}: Treatment pathways are step-by-step plans outlining the recommended medical
    care for specific diseases; they get revised when different treatments are found to improve
    patient outcomes. Examining health records is an important part of this revision process, but
    inferring patients' actual treatments from health data is challenging due to complex
    event-coding schemes and the absence of pathway-related annotations. The objective of this study
    is to develop a method for inferring actual treatment steps for a particular patient group from
    administrative health records — a common form of tabular healthcare data — and address several
    technique- and methodology-based gaps in treatment pathway-inference research. 

    \emph{Methods}: We introduce \emph{Defrag}, a method for examining health records to infer the
    real-world treatment steps for a particular patient group. \emph{Defrag} learns the semantic and
    temporal meaning of healthcare event sequences, allowing it to reliably infer treatment steps
    from complex healthcare data. To our knowledge, \emph{Defrag} is the first pathway-inference
    method to utilise a neural network (NN), an approach made possible by a novel, self-supervised
    learning objective. We also developed a testing and validation framework for pathway inference,
    which we use to characterise and evaluate \emph{Defrag}'s pathway inference ability, establish
    benchmarks, and compare against baselines.

    \emph{Results}: We demonstrate \emph{Defrag}'s effectiveness by identifying best-practice
    pathway fragments for breast cancer, lung cancer, and melanoma in public healthcare records.
    Additionally, we use synthetic data experiments to demonstrate the characteristics of the
    \emph{Defrag} inference method, and to compare \emph{Defrag} to several baselines, where it
    significantly outperforms non-NN-based methods.
    
    \emph{Conclusions}: \emph{Defrag} offers an innovative and effective approach for inferring
    treatment pathways from complex health data. \emph{Defrag} significantly outperforms several
    existing pathway-inference methods, but computationally-derived treatment pathways are still
    difficult to compare against clinical guidelines. Furthermore, the open-source code for
    \emph{Defrag} and the testing framework are provided to encourage further research in this area.
\end{abstract}




\maketitle

\input{paper.tex}

\section*{CRediT authorship contribution statement}

\textbf{A. Wilkins-Caruana}: Conceptualisation, methodology, software, validation, formal analysis,
investigation, resources, data curation, writing - original draft, visualisation.

\textbf{M. Bandara, K. Musial}: Writing - review and editing, supervision.

\textbf{D. Catchpoole, P. J. Kennedy}: Writing - review and editing, supervision, project administration,
funding acquisition.

\section*{Declaration of Competing Interest} 

The authors declare that they have no known competing financial interests or personal relationships
that could have appeared to influence the work reported in this paper.

\section*{Acknowledgements}

Funding: This work and the PhD scholarship of Adrian Wilkins-Caruana was supported through a national initiative by Cancer Australia as part of an approach to improving national cancer data on stage, treatment and recurrence.

\bibliographystyle{cas-model2-names}

\bibliography{ref}

\appendix

\input{appendix.tex}
\end{document}

%% file: paper.tex
\hypertarget{introduction}{%
\section{Introduction}\label{introduction}}

Treatment pathways are step-by-step plans outlining the recommended
interventions for treating a group of patients with a particular disease
\citep{Rotter_2011}. For example, a breast cancer treatment pathway may
specify suitable interventions, such as surgery or chemotherapy, and
their order, like pre- or post-surgical chemotherapy
\citep{Cardoso_2019}. Treatment pathways are a central component of
clinical pathway research, a discipline that spans a patient's entire
healthcare journey, from initial screening to end-of-life care.
Treatment pathways, which specifically address therapeutic
interventions, minimise variability in clinical practice, and ultimately
lead to improved patient outcomes
\citep{Panella_2003,Rotter_2010,Rotter_2011}.

In practice, an individual patient's actual treatments can deviate from
best-practice treatment pathways due to demographic, geographic, and
socioeconomic factors \cite{DiMatteo_2004,Ebben_2013}. Examining the
health records of a patient cohort to determine their actual pathway of
treatments offers valuable insights into treatment patterns that can
lead to improved treatment pathways \cite{Fauman_2007,Yu_2014}. For
instance, identifying the most effective cancer treatments can shape
treatment pathway recommendations. However, using health records to
develop treatment pathways is challenging due to the interplay of many
clinical factors \cite{Chen_2017,Yadav_2018}. Issues such as patient
traits (e.g., comorbidities or genetic predispositions) and
environmental factors (e.g., resource availability or practitioner
experience) create inconsistencies in health records, making identifying
the cohort's actual treatment pathway tough.

To address these challenges, techniques such as process mining
\cite{De_Weerdt_2013,Baker_2017,Litchfield_2018,Lim_2021,Lim_2022} and
probabilistic models
\cite{Zhang_2015,Zhang_2015a,Huang_2013,Huang_2014,Xu_2016,Xu_2017,Huang_2018}
are commonly used to identify pathway-related concepts in healthcare
data. Process mining studies primarily focus on identifying simpler
structures like common event sequences (e.g., event B follows event A),
but struggle to capture dynamic and unstructured medical processes
\cite{Yang_2014}. Probabilistic methods are capable of inferring more
complex patterns (e.g., event B follows event A with a given
probability); however, they face limitations in addressing long-range
temporal dependencies, high dimensionality, rare or noisy events, and
may not accurately portray the influence of past events on future
decisions.

This study aims to enhance treatment pathway inference research by
introducing a new method, \emph{Defrag}, to infer actual treatment
pathways from administrative health records (AHRs), a common form of
tabular healthcare data. \emph{Defrag} comprises a Transformer neural
network (NN) that is trained on AHRs using a novel, self-supervised
training objective called the
\emph{semantic-temporal learning objective} (\emph{STLO}). \emph{Defrag}
infers pathways in two steps: 1) the Transformer generates a joint
semantic-temporal encoding of AHR events, and 2) \emph{Defrag} applies
network inference to determine sequential correlations between clusters
of encoded events, uncovering actual treatment pathways. \emph{Defrag}
excels at capturing complex, long-range dependencies, modelling
non-linear relationships, learning context-aware event representations,
and demonstrating robustness to noise. Using self-supervised learning
without prior pathway-specific labels or pre-trained encodings,
\emph{Defrag} can adapt to various healthcare settings, accommodate data
variations across regions, and remain relevant as medical practices
change.

We also present a technique for generating synthetic AHRs, which we
utilise for testing pathway inference methods, including \emph{Defrag}.
Synthesising AHRs from ground-truth pathways enables quantitative
evaluation of \emph{Defrag}'s effectiveness in inferring actual
treatment pathways. It helps characterise the strengths and weaknesses
of \emph{Defrag} while also allowing benchmarking and comparisons with
alternative approaches. This technique also helps assess \emph{Defrag}'s
performance in relation to various data factors, such as treatment
consistency, code-set vocabulary size, and pathway complexity. This
approach is currently the only method for quantitatively evaluating
pathway inference methods.

We demonstrate \emph{Defrag} by identifying fragments of best-practice
treatment pathways for breast cancer \cite{Cardoso_2019}, lung cancer
\cite{Vansteenkiste_2014}, and melanoma \cite{Michielin_2019} from the
publicly available MIMIC-IV dataset \cite{Johnson_2020}. Furthermore, we
show in synthetic experiments that \emph{Defrag} is substantially more
accurate than several baselines at inferring synthetic pathways. We also
open-source our code to facilitate the reproducibility of our results
and research iteration\footnote{Code is available at:
\url{https://github.com/adriancaruana/defrag}}.

We summarise the significance of this paper as follows:

\textbf{Problem}: Inferring actual treatment pathways from complex
administrative health records (AHRs) is challenging due to variations in
patient comorbidities, responses, and resource availability.

\textbf{What is Already Known}: Existing methods struggle with capturing
dynamic and unstructured medical processes, hindering pathway inference.
Limited open-source data and evaluation techniques further impede
reproducibility and validation.

\textbf{What this Paper Adds}: We introduce Defrag, a novel method
utilising a Transformer neural network for inferring treatments in AHRs.
We also present a novel method for pathway inference validation using
synthetic data. Defrag significantly outperforms existing methods and
identifies best-practice treatment patterns for three cancer types in
public AHRs.

\hypertarget{background-and-related-work}{%
\section{Background and Related
Work}\label{background-and-related-work}}

Administrative health records (AHRs) are temporal, tabular datasets
collected for administrative purposes, such as insurance and billing.
These records capture event logs of specific treatment activities, such
as diagnoses and procedures. However, numerous clinical factors make
pathway inference challenging \cite{Chen_2017,Yadav_2018}. Such factors
include a patient's comorbidities, genetic predispositions, and
treatment responses, as well as other healthcare factors such as
resource availability, practitioner experience, and geographic,
socioeconomic, and cultural factors.

Process mining has been investigated for discovering treatment pathways
in AHRs
\cite{De_Weerdt_2013,Baker_2017,Litchfield_2018,Lim_2021,Lim_2022}.
Process mining is a method of uncovering rule-based patterns in event
logs which assumes the underlying process occurs in a structured manner
\cite{Van_Der_Aalst_2012}. This assumption is a known limitation of
process mining in AHRs, particularly for human-centric clinical pathways
\cite{Guzzo_2021}. AHRs often produce unstructured patterns driven by
the decisions of individual patients and healthcare practitioners rather
than strict business processes \cite{Lang_2008,Rebuge_2012,Yang_2014}.
Other studies have applied related techniques to directly infer temporal
patterns between events in AHRs for heart failure and EHR workflow
patterns \cite{Yan_2016,Zhang_2022}. These approaches struggle to
account for semantically similar events or long-range temporal
dependencies, which are essential components of treatment pathways, and
thus these factors limit the applicability of process mining for pathway
inference \cite{Yang_2014,Guzzo_2021}.

Probabilistic models such as Markov chains
\cite{Zhang_2015,Zhang_2015a}, Latent Dirichlet Allocation (LDA) or
other topic models \cite{Huang_2013,Huang_2014,Xu_2016,Xu_2017}, and
Hidden Markov Models \cite{Huang_2018} may better identify clinical
pathways in unstructured, human-centric AHRs. Unlike process mining,
these methods can capture more abstract processes. However, they often
oversimplify AHRs (e.g., removing the temporal component) or impose
model constraints (e.g., the Markov condition) \cite{Ghahramini_2001},
limiting their ability to discover non-linear temporal patterns or
account for the impact of past events on future decisions in
unstructured and noisy AHRs.

The need for open-source data and code limits AHR-based clinical pathway
research. Except for one study \cite{De_Weerdt_2013}, all research
relies on closed-source datasets, and none share their code. Moreover,
only some studies validate data-derived clinical pathways against
real-world pathways, which is challenging due to the semantic gap
between clinically-developed and computationally-derived treatment
pathways. Furthermore, AHRs do not capture clinical pathway-related
concepts, and clinical pathways need to be computer-interpretable,
making inferred pathways difficult to validate \cite{Oliart_2022}.

In this study, we apply sequence-based neural networks (NNs) to model
AHRs for pathway inference. NN-based methods (e.g., Word2Vec
\cite{Mikolov_2013} and Transformers \cite{Vaswani_2017}) are highly
applicable for processing natural language, a form of data that
resembles the unstructured sequences of treatment events typically found
in AHRs. NN-based methods have been applied successfully to AHR analysis
(e.g., Med2Vec \cite{Choi_2016}, and MiME \cite{Choi_2018}), but they
have yet to be used for inferring treatment pathways.

To enhance the accessibility of pathway inference methods, we offer
open-source code for our methodology. Additionally, we develop a
synthetic data generation method to facilitate performance evaluation
and comparison of pathway inference methods. We also demonstrate results
on the publicly available MIMIC-IV AHR dataset \cite{Johnson_2020}.

\hypertarget{method}{%
\section{Method}\label{method}}

This section outlines our methodological contributions in two parts.
Section \ref{pathway-defragmentation-method} pertains to the pathway
defragmentation approach and Section
\ref{testing-and-validation-framework} pertains to synthetic AHR data
generation.

\hypertarget{pathway-defragmentation-method}{%
\subsection{Pathway Defragmentation
Method}\label{pathway-defragmentation-method}}

\begin{figure*}[t]
    \centering
    \includegraphics[width=\linewidth]{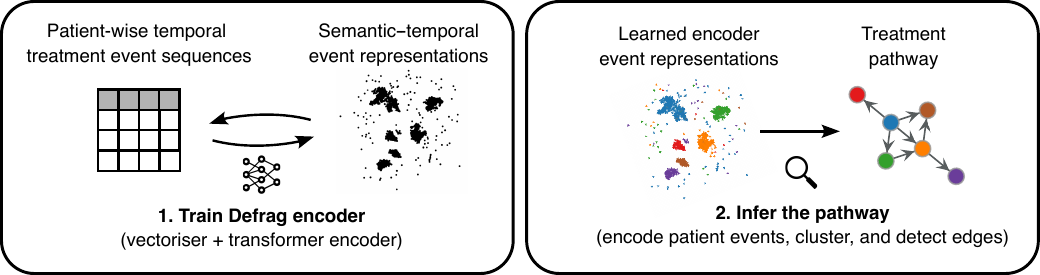}
    \caption{
      \textbf{Pathway defragmentation} (\textit{Defrag}) is a 2-step process for inferring
      actual treatment pathways from administrative health records (AHR). For the first step, we use
      patients' health records to train an event encoder that maps categorical treatment
      event codes to a semantic-temporal numerical representation via a Transformer neural
      network and a novel self-supervised learning objective. In the second step, we encode the patients'
      treatment events, cluster the encoded event representations, and infer a graph of the treatment
      clusters using the temporal information in the raw event sequences. 
    }
    \vspace{-5mm}
    \label{fig:defrag_main}
\end{figure*}

We introduce \emph{Defrag}, a novel technique for identifying treatment
pathways in AHRs. \emph{Defrag} comprises two components, as illustrated
in Figure \ref{fig:defrag_main}. The first component generates a
temporally contextualised encoding of AHR events, learning patterns of
events frequently occurring in close temporal proximity across numerous
patients, resulting in a mapping that preserves temporal closeness. For
instance, though distinct, anaesthetic and endoscopy procedures should
have similarly encoded representations, as they are often utilised
concurrently.

The second component infers a pathway by partitioning the encoded event
space into clusters, identifying macro-scale temporal relationships
between partitions. An example would be the frequent occurrence of
diagnostic procedures before therapeutic procedures. The clustering
groups similar events, and the temporal relationships contribute to the
pathway graph formation.

\hypertarget{ahr-data-representation}{%
\subsubsection{AHR Data Representation}\label{ahr-data-representation}}

\begin{figure*}
  \center{
    \includegraphics[width=\linewidth]{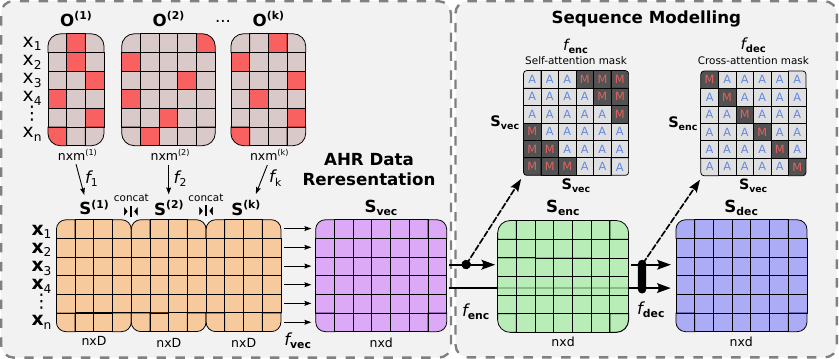}
  }
  \vspace{-2mm}
  \caption{
    The flow of AHR events' transformation into encoded/decoded vector representations. Categorical
    variables are one-hot encoded (red), then mapped to fixed-length vectors by
    $f_k$ and concatenation (orange), then to a vector of size $d$ by
    $f_\mathbf{vec}$ to produce $\mathbf{S_{vec}}$ (purple). Then, the transformer encoder
    $f_{\mathbf{enc}}$ maps $\mathbf{S_{vec}}$ to $\mathbf{S_{enc}}$ (green), and finally the transformer
    decoder $f_{\mathbf{dec}}$ maps $\mathbf{S_{vec}}$ and $\mathbf{S_{enc}}$ to $\mathbf{S_{dec}}$
    (blue). The attention masks in the encoder and decoder for $n=6$ events are depicted in the
    upper-right. $w=2$ for the encoder's windowed mask. ``A'' and ``M'' indicate positions which can
    and cannot attend respectively.  
  }
  \vspace{-4mm}
  \label{fig:vector_shapes_flow}
\end{figure*}

AHRs are typically tabular, with rows \(i \in \{1, \dots, n\}\)
representing different events over time, and columns
\({X^{(1)}, \dots, X^{(k)}}\) for different variables (e.g., diagnosis
codes and procedure codes). We seek to represent \(n\) events
observations from an arbitrary number of variables \(k\) as a single
matrix. This process is shown in Figure \ref{fig:vector_shapes_flow}
(left).

For any patient, consider the sequence
\(x^{(k)}_{1}, x^{(k)}_{2}, \dots, x^{(k)}_{n}\) of \(n\) events (rows)
from a single categorical variable \(X^{(k)}\). If the cardinality of
\(X^{(k)}\) is \(m^{(k)}\) , then the sequence can be represented as an
\(n \times m^{(k)}\) matrix \(\mathbf{O}^{(k)}\) of \(n\) one-hot
vectors. Additionally, consider several variables
\(X^{(1)}, X^{(2)}, \dots, X^{(k)}\); we represent a sequence of each
variable as an \(n \times m^{(k)}\) matrix similarly.

For each variable \(X^{(1)}, X^{(2)}, \dots, X^{(k)}\), we use
multilayer perceptrons to learn mappings \(f_1, f_2, \dots, f_k\) from
the one-hot encoded vectors to vectors of size \(D\), which are applied
event-wise to
\(\mathbf{O}^{(1)}, \mathbf{O}^{(2)}, \dots, \mathbf{O}^{(k)}\) to yield
\(n\times D\) shaped sequence matrices for each variable, denoted as
\(\mathbf{S}^{(1)}, \mathbf{S}^{(2)}, \dots, \mathbf{S}^{(k)}\). By
concatenating the vectors for each event in the sequence along the
columns, \(n\) observations from several variables \(k\) can be
represented as a single \(n\times kD\) matrix. While numerical features
are less common in AHRs, they too can be included in the concatenation.
Finally, another event-wise mapping \(f_\mathbf{vec}\) reduces the
dimension of the concatenated vectors (length \(kD\)) to \(d\). The
result is a single sequence matrix of event vectors \(\mathbf{S_{vec}}\)
of shape \(n\times d\), which numerically represents all \(n\)
observations from all \(k\) variables.

The benefit of this approach is threefold: 1) it significantly reduces
the dimensionality of the one-hot encoded features, reducing their
redundancy and mitigating the computational burden of the subsequent
sequence model (see Section \ref{sequence-modelling}), 2) it enables the
conversion of categorical data common in AHRs to numerical data required
by NNs, and 3) it does not preclude numerical features or pre-trained
encodings (e.g., Med2Vec \cite{Choi_2016}) from being utilised, which
can also be included in the \(f_\mathbf{vec}\) mapping.

\hypertarget{sequence-modelling}{%
\subsubsection{Sequence Modelling}\label{sequence-modelling}}

The vectorised events in \(\mathbf{S_{vec}}\) only contain semantic
information, e.g., a vector representing anaesthesia use in the ICU.
This section discusses how to create new vector representations of
events that contain both semantic \emph{and} temporal context. e.g., a
patient is anaesthetised in the ICU for an endoscopy. We use the
semantic event vectors \(\mathbf{X}_{i, \mathbf{vec}}\) in
\(\mathbf{S_{vec}}\) to learn new event vectors that contain semantic
\emph{and} temporal information using a Transformer encoder-decoder
architecture \cite{Vaswani_2017}. The encoder \(f_{\mathbf{enc}}\) and
decoder \(f_{\mathbf{dec}}\) are trained to generate new \(d\)-shaped
vectors for all \(n\) events in \(\mathbf{S_{vec}}\).

We diverge from the original Transformer configuration in
\cite{Vaswani_2017} in three ways. 1) \emph{Defrag}'s Transformer is
trained with a unique, self-supervised objective that emphasises
learning temporal features (Section \ref{training-and-loss-function}).
2) We \emph{relative} (instead of \emph{absolute}) positional encoding
\cite{Shaw_2018, Huang_2018b}. 3) We adjust the attention masks in the
encoder and decoder. Specifically, the encoder uses windowed
self-attention mask in the encoder, with each event attending to \(w\)
of its neighbours on either side. A lower \(w\) leads to more
event-level encodings, while a higher \(w\) results in more
patient-level encoding. We also mask diagonal positions in the decoder's
self- and cross-attention masks to prevent information from flowing
directly between the model's input and output and to promote encoder and
decoder to learn distinct features. The attention masks are depicted in
the upper-right corner of Figure \ref{fig:vector_shapes_flow}, and
Appendix \ref{windowed-attention} analyses the implications of these
decisions, including the effect of \(w\) on the learned event
representations.

\hypertarget{training-and-loss-function}{%
\subsubsection{Training and Loss
Function}\label{training-and-loss-function}}

We train the parameters \(\theta\) of \emph{Defrag}'s Transformer to
generate joint semantic-temporal representations of events. Supervised
learning is not possible since AHRs contain no semantic-temporal labels
nor any other treatment pathway-specific attributes or annotations.
Instead, we explored several popular self-supervised learning objectives
-- e.g., auto-encoding, Barlow-twins \cite{Zbontar_2021}, SimCSE
\cite{Gao_2021} -- but all were suboptimal for pathway inference (see
Appendix \ref{stlo-discussion}).

\begin{figure}[t]
  \center{
    \includegraphics[width=\linewidth]{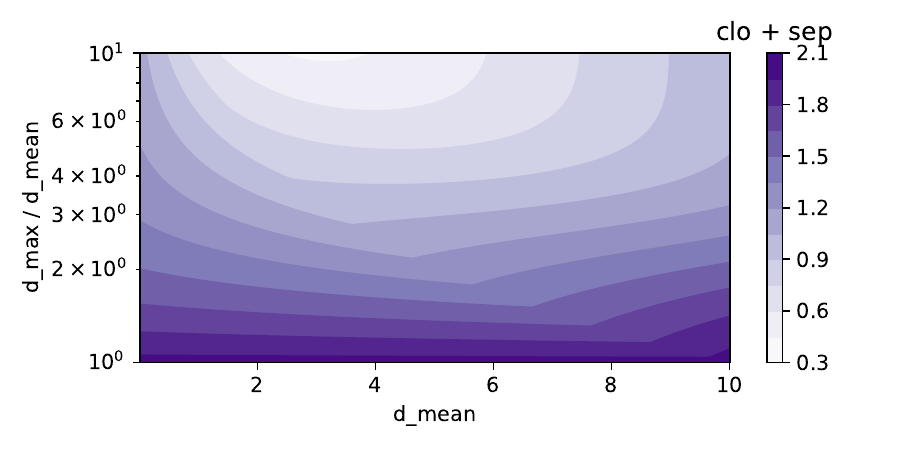}
  }
  \vspace{-2mm}
  \caption{
    A contour plot showing the loss landscape for the sum of the $clo$ and $sep$ components of
    $\mathcal{L}_{\mathrm{STLO}}$, which vary for different values of $d_{\mathrm{mean}}$ and
    $d_{\mathrm{max}}$. 
  }
  \vspace{-2mm}
  \label{fig:loss_function}
\end{figure}

We designed a novel objective called the
\textit{semantic-temporal learning objective} (STLO) to optimise
\(\theta\). STLO is a contrastive objective that encourages the
Transformer to learn encoded representations of events that expose their
semantic and temporal qualities . Such information is critical for
pathway inference since some events in the sequence will be related to
the same treatment (e.g., procedure and anaesthetic events during
surgery), while some events in the sequence might be a part of different
treatments (e.g., surgery and then chemotherapy).

\(\mathcal{L}_{\mathrm{STLO}}\) consists of three components:
\emph{closeness} (\(clo\)), \emph{separation} (\(sep\)), and
\emph{consistency} (\(con\)). These components are used to separate a
sequence of events into \emph{two} temporally contiguous groups.
\emph{Clo} (Equation \ref{eq:clo}) measures the within-group distances
-- it is minimised when the representations of such events are similar.
\emph{Sep} (Equation \ref{eq:sep}) measures the between-group distance
-- it is minimised when the representations between the two groups are
dissimilar. Finally, \emph{con} (Equation \ref{eq:con}) can be
interpreted as a regularisation term -- it is minimised when the
distribution of representations across groups does not vary
significantly.

\(\mathcal{L}_{\mathrm{STLO}}\) is computed as follows: for
\(\mathbf{S_{rep}}\), which can be either \(\mathbf{S_{enc}}\) or
\(\mathbf{S_{dec}}\), the sequence of \(n\) event vectors is used to
compute a sequence of \(n-1\) Euclidean distances
\(\mathbf{dist}_{\mathrm{rep}}\) between adjacent representations. From
these distances, we compute \(\mathrm{dist}_{\mathrm{mean}}\) and
\(\mathrm{dist}_{\mathrm{max}}\) as follows: \begin{align}
 \mathrm{dist}_{\mathrm{mean}} &= \frac{1}{n-1}\sum_{i=1, i \neq \argmax{\mathbf{dist}}}^{n-1}(\mathbf{dist}_{\mathrm{rep}, i}), \\
 \mathrm{dist}_{\mathrm{max}} &= \max_{i=1}^{n-1}{\left(\mathbf{dist}_{\mathrm{rep}, i}\right)}.
\end{align} where
\(0 < \mathrm{dist}_{\mathrm{mean}} \leq \mathrm{dist}_{\mathrm{max}}\).
Then, we compute \(clo\), \(sep\) and \(con\): \begin{align}
 clo &= \mathrm{dist}_{\mathrm{mean}} / \min{(\mathrm{dist}_{\mathrm{max}} + \epsilon, \zeta)}  \label{eq:clo}\\
 sep &= 1 - \tanh{\left(\left(\mathrm{dist}_{\mathrm{max}} - \mathrm{dist}_{\mathrm{mean}}
 \right)/\eta\right)} \label{eq:sep} \\
 con &= \std_{i=1, i \neq \argmax{\mathbf{dist}}}^{n-1}{(\mathbf{dist}_{\mathrm{rep}, i})} \label{eq:con}
\end{align} where \(\zeta\) and \(\eta\) are constants which are used to
equilibrate the scales of \(clo\) and \(sep\), and \(\epsilon\) is a
small positive value to prevent division by zero. We find
\(\zeta=10, \eta=20\), and \(\epsilon=10^{-8}\) to be suitable for all
our experiments. We choose these specific formulations because they are
simple to compute and effective in practice . To combine these
components, we compute \(\mathcal{L}_{\mathrm{STLO}}\) on
\(\mathbf{S_{enc}}\) and \(\mathbf{S_{dec}}\) separately, and take the
loss at each training step as
\(clo_\mathbf{dec} + sep_\mathbf{dec} + con_\mathbf{enc}\). Figure
\ref{fig:loss_function} shows how minimising \(clo + sep\) both
minimises the within-group distances \(\mathrm{dist}_{\mathrm{mean}}\),
while simultaneously maximising the between-group distance
\(\mathrm{dist}_{\mathrm{max}}\), thus separating the sequence into two
groups.

During training, contiguous segments from patients' event logs are
sampled as sequences of length \(n\). Sequences shorter than \(n\)
events are padded to the right, but the STLO is computed only on the
valid portion of each sequence.

Notably, \(\mathcal{L}_{\mathrm{STLO}}\) computes \(clo\) and \(sep\)
using \emph{Defrag}'s decoded sequence representation
\(\mathbf{S_{dec}}\), while the semantic-temporal representations used
for pathway inference are the encoder's output \(\mathbf{S_{enc}}\). We
highlight that \(\mathbf{S_{dec}}\) is \emph{not} used for pathway
inference since STLO's inductive biases may not lead to useful
representations. For example, sequences containing more or fewer than
two kinds of treatment groups would be valid yet incompatible with STLO.
However, these inductive biases are \emph{not} placed on the encoder; it
too must learn both semantic and temporal information since the decoder
cannot rely on such information from the source sequence due to its
attention mask (as discussed in Section \ref{sequence-modelling}).

Therefore, the role of \(\mathbf{S_{dec}}\) is to facilitate learning of
semantic-temporal representations in \(\mathbf{S_{enc}}\) via
\(\mathcal{L}_{\mathrm{STLO}}\) despite its inductive biases. We provide
further details regarding the design of the STLO and comparisons with
alternative self-supervised learning objectives in Appendix
\ref{stlo-discussion}.

\hypertarget{pathway-inference}{%
\subsubsection{Pathway Inference}\label{pathway-inference}}

Pathway inference, which refers to inferring a directed graph \(G\) from
\(\mathbf{X}_{\mathbf{enc}}\), is achieved in four main steps: encoding,
clustering, inference, and post-processing. We describe these steps
below.

\textbf{Encoding}: The AHR events \(X_{i}\) in a patient's event
sequence \(S\) are encoded with a trained mapping
\(f_{\mathbf{enc}} \cdot f_{\mathbf{vec}} \cdot f_{k}\) to obtain the
semantic-temporal representations \(\mathbf{S_{enc}}\). Note, while the
NN is trained with sequences of fixed length \(n\), inference can be
performed on sequences of any length (i.e., the patient's entire event
sequence). For example, the \(k\) features from a patient's event log of
\(l\) events are first encoded with each categorical mapping \(f_k\),
then each mapping is merged via \(f_{\mathbf{vec}}\), and finally, the
\(l\) merged mappings are encoded with \(f_{\mathbf{enc}}\). The event
sequences of individual patients in a cohort are each encoded separately
using the same trained mapping, yielding \(\mathbf{X}_{\mathbf{enc}}\),
a concatenated matrix of sequences \(\mathbf{S_{enc}}\) for all
patients.

\textbf{Clustering}: The vertices in \(G\) are obtained by clustering
\(C(\mathbf{X}_{\mathbf{enc}})\). We find that hierarchical algorithms
like agglomerative clustering or HDBSCAN \cite{McInnes_2017} typically
work best; however, the choice of clustering algorithm largely depends
on the shape and density of \(\mathbf{X}_{\mathbf{enc}}\). Optionally,
the clustering can be optimised using a parameter search to optimise an
unsupervised clustering metric (e.g., Calinski-Harabasz
\cite{Calinski_1974}). Appendix \ref{cluster-optimisation} provides
further details on the clustering algorithm and parameter search.

\textbf{Inference}: We construct a directed graph \(G = (V, E)\) by
interpreting the clusters in \(C(\mathbf{X}_{\mathbf{enc}})\) as
vertices \(V\) and establishing weighted, directed edges \(E\) between
vertices when adjacent events in a patient's sequence belong to
different clusters, denoting a ``from \(\to\) to'' edge. The cluster of
an event \(X_{i}\) in a patient's event sequence is given as
\(C(\mathbf{X}_{\mathbf{enc},i})\), and the weight \(w\) of the directed
edge from vertex \(p \in V\) to vertex \(q \in V\) is given by:
\begin{equation} 
\label{eq:defrag-infer} 
w_{p, q, p \neq q} =
\sum_{i=1}^{n-1}{ \begin{cases} 1, & \mathrm{if}~C(\mathbf{X}_{\mathbf{enc},i}) = p~\mathrm{and}~C(\mathbf{X}_{\mathbf{enc},{i+1}}) = q \\
        0 & \mathrm{otherwise}
    \end{cases}
}
\end{equation} where \(i\) enumerates the events in a patient's
treatment sequence. Equation (\ref{eq:defrag-infer}) is applied
per-patient, aggregated by summing the weights for all patients.
Furthermore, Equation (\ref{eq:defrag-infer}) is applied to all pairwise
combinations of vertices to determine the weights between all vertices
in \(G\).

\textbf{Post-processing}: The edge weights \(w\) yield a graph with
bidirectional edges \(G_{\leftrightarrow}\). A simplified graph with
unidirectional edges \(G_{\to}\) is obtained by collapsing bidirectional
edges to the more strongly-weighted direction. The adjacency matrix
\(A_{\to}\) of \(G_{\to}\) is defined as
\(A_{\to} = A_{\leftrightarrow} - A_{\leftrightarrow}^T\), where
\(A_{\leftrightarrow}\) is the adjacency matrix of
\(G_{\leftrightarrow}\). The edge weights are normalised using
\(A / \max(A)\), and an optional threshold can be used to binarise the
edges. \(G_{\to}\) is useful for observing the salient relationships in
\(G\), while \(G_{\leftrightarrow}\) is more useful for assessing
individual relationships of interest, such as examining the specific
weight and direction between two particular vertices.

\hypertarget{testing-and-validation-framework}{%
\subsection{Testing and Validation
Framework}\label{testing-and-validation-framework}}

\emph{Defrag} cannot be quantitatively validated against evidence-based
treatment pathways because AHRs contain no pathway-specific annotations.
Instead, we validate \emph{Defrag} by inferring synthetic pathways,
which are used to generate synthetic AHRs. This approach offers three
advantages: 1) it allows us to vary the AHR distribution and complexity
by configuring the synthetic data generation parameters, 2) it enables
us to quantify \emph{Defrag}'s efficacy by comparing inferred and
synthetic pathways, and 3) it can be used to characterise
\emph{Defrag}'s strengths and weaknesses across various synthetic
datasets.

Our synthetic data generator requirements differ from other synthetic
AHR generation methods in several ways. The traditional goal of
synthetic AHR generators is to reproduce variable distributions and
inter-variable relationships in existing AHRs without disclosing
sensitive subject information \cite{Goncalves_2020}; however, we aim to
test pathway inference methods, such as \emph{Defrag}, for identifying
pathways and temporal relationships in tabular data. Furthermore, AHRs
typically do not capture treatment pathway-related concepts; thus, we
cannot generate synthetic AHRs with traditional methods without
pathway-related annotations. Finally, because we do not intend to
replicate AHRs for redistribution, our synthetic AHRs do not need to
align as closely as possible to sampled AHRs.

For these reasons, we design a generative algorithm that produces
plausible AHRs based on a randomly generated graph. The synthetic data
generation begins by generating a random directed graph \(G_{syn}\),
which serves as a ground-truth pathway. Vertices \(v\) in
\(G_{\mathrm{syn}}\) which have no outward-directed edges are ``end''
vertices \(v^{\mathrm{end}}\), and ones with no inward-directed edges
are ``start'' vertices \(v^{\mathrm{start}}\). We generate patient
pathways as random walks through \(G_{\mathrm{syn}}\). We use a directed
version of the extended Barabási--Albert model graph
\cite{Albert_2000}\footnote{We
modify the undirected
\href{https://networkx.org/documentation/stable/reference/generated/networkx.generators.random_graphs.extended_barabasi_albert_graph.html}{networkx}
implementation to generate directed graphs, with edges directed outward from earlier-generated
vertices.} for all our experiments. We chose this model since it
generates graphs with desirable properties such as branching,
non-uniform edge frequency in random walks, and possible but infrequent
cycles.

We use \(G_{syn}\) to generate a sequence of vertices (random walk) and
events for each patient. At each index in the sequence, a vertex \(v_i\)
and an event \(X_i\) are sampled, as well as a vertex advancement
probability
\(p_{i, \mathrm{adv}} \leftarrow \mathrm{Bernoulli}(\delta)\), where
\(\delta\) controls the transition likelihood. The first vertex is
sampled uniformly from the set of starting vertices, while subsequent
vertices are sampled as follows: \begin{equation}
  v_i = 
  \begin{cases} 
    \mathrm{Uniform}(v_{\mathrm{adv}})  & p_{i, \mathrm{adv}} = 1 \\
                    v_{i-1}            & p_{i, \mathrm{adv}} = 0, \\
  \end{cases}
\end{equation} where \(v_{\mathrm{adv}}\) denotes the set of
outward-directed edges from \(v\). If \(p_{i, \mathrm{adv}} = 1\) and
\(v_{\mathrm{adv}} = \varnothing\), then \(i\) is the last index of the
sequence.

Events are sampled from the discrete random variable \(X\) as follows:
\begin{align}
  X_i &= P(X \mid v_i) \sim \mathrm{Zipf}(a_{v_{i}}), \\
  \mathrm{supp}(P) &= \mathrm{Perm}(\mathbb{X})
\end{align} where \(\mathrm{Perm}(\mathbb{X})\) denotes a random
permutation over the set of possible values of \(X\), and \(a\) is the
\(\mathrm{Zipf}\) distribution exponent. We choose to model the
distribution of events using the \(\mathrm{Zipf}\) distribution since,
to a good first approximation, the \(n^\mathrm{th}\) most common event
in a large AHR corpus appears with frequency \(1/n\); this property is
shared with the word frequency in natural language, and other forms of
physical and social data \cite{Yang_2020,Piantadosi_2014,Wolfram_2002}.
Furthermore, since treatment-related concepts are not captured in AHRs,
it is not possible to directly model \(P(X \mid v_i)\). We show that
this method yields plausible tabular AHRs in Appendix
\ref{synthetic-ahr-plausibility} by comparing the distribution of
generated and authentic AHR events.

\hypertarget{experiments-and-results}{%
\section{Experiments and Results}\label{experiments-and-results}}

This section explores \emph{Defrag}'s pathway inference ability. As a
case study, we first use MIMIC-IV \cite{Johnson_2020} to identify cancer
treatment pathways. Next, we use the testing and validation framework
from Section \ref{testing-and-validation-framework} to quantify
\emph{Defrag}'s pathway inference performance. We train all models with
32GB RAM, 6-core Intel i7 CPU, and NVIDIA 10-series GPU with 8GB VRAM.

\hypertarget{mimic-iv-experiments}{%
\subsection{MIMIC-IV Experiments}\label{mimic-iv-experiments}}

We apply Defrag to identify cancer treatment pathway fragments in the
MIMIC-IV dataset \cite{Johnson_2020}, the largest public AHR dataset. We
analyse treatment pathways in cancer cohorts, which are well-suited for
pathway inference, as the availability of established treatment pathways
from the European Society for Medical Oncology (ESMO) allows for an
empirical comparison with data-driven inferences. While MIMIC-IV is not
cancer-specific, it contains clinical data from over 40,000 ICU
admissions, many of which are from cancer patients. We focus on cancers
with a typical surgical component due to the ICU context of MIMIC-IV:
breast cancer, lung cancer, and melanoma. Patient cohorts are defined by
\href{https://www.hcup-us.ahrq.gov/toolssoftware/ccs/ccs.jsp}{Clinical Classifications Software}
(CCS) cancer diagnosis categories. Table \ref{table:mimic_data_info}
shows each dataset's number of patients, hospital admissions, and unique
procedure codes.

\begin{table}
    \caption{A summary of each of the MIMIC-IV cancer datasets.}
    \label{table:mimic_data_info}
    \center{
      \begin{tabular}{lrrrr}
          \toprule
          Experiment & Events & Patients & Admissions & Codes \\
          \midrule
          breast & 14178 & 1576 & 3926 & 1139 \\
          lung & 10132 & 958 & 2483 & 787 \\
          melanoma & 4019 & 482 & 1221 & 689 \\
          \bottomrule
      \end{tabular}
    }
\end{table}

Each experiment uses \(d=128\), \(16\) attention heads, \(8\) encoder
and decoder layers, a feedforward dimension of \(64\), and a dropout of
\(0.2\). Each model is trained for \(50k\) steps with a batch size of
\(128\) and optimised using AdamW \cite{Loshchilov_2019} with a learning
rate of \(10^{-4}\). These parameters were determined empirically
through extensive experimentation, ensuring consistent performance
across each dataset while balancing model complexity and computational
efficiency. Events are multivariate features consisting of the hospital
admission index of the patient, the ICD9 code, and three levels of the
ICD9 code's hierarchical CCS categorisation. We encode missing values as
all-zero in the one-hot encoding. Since MIMIC-IV contains both ICD9 and
ICD10 codes, we map each ICD10 code to ICD9 for
consistency.\footnote{We use the general equivalence mapping from
\href{https://www.nber.org/research/data/icd-9-cm-and-icd-10-cm-and-icd-10-pcs-crosswalk-or-general-equivalence-mappings}{Centers
for the Medicare \& Medicaid Services} to convert between ICD 9 and 10.}

After training \emph{Defrag} and generating event representations, we
use hierarchical clustering with five clusters to classify treatments.
Appendix \ref{cluster-optimisation} provides further clustering choice
details. We also generate TF-IDF weighted histograms of procedure code
frequencies for each cluster based on the original patient sequences to
highlight the important codes in each cluster and use the CCS ontology
to hierarchically group similar codes for increased interpretability.

\begin{figure}[t]
    \begin{subfigure}{\columnwidth}
        \center{
          \includegraphics[width=0.95\linewidth]{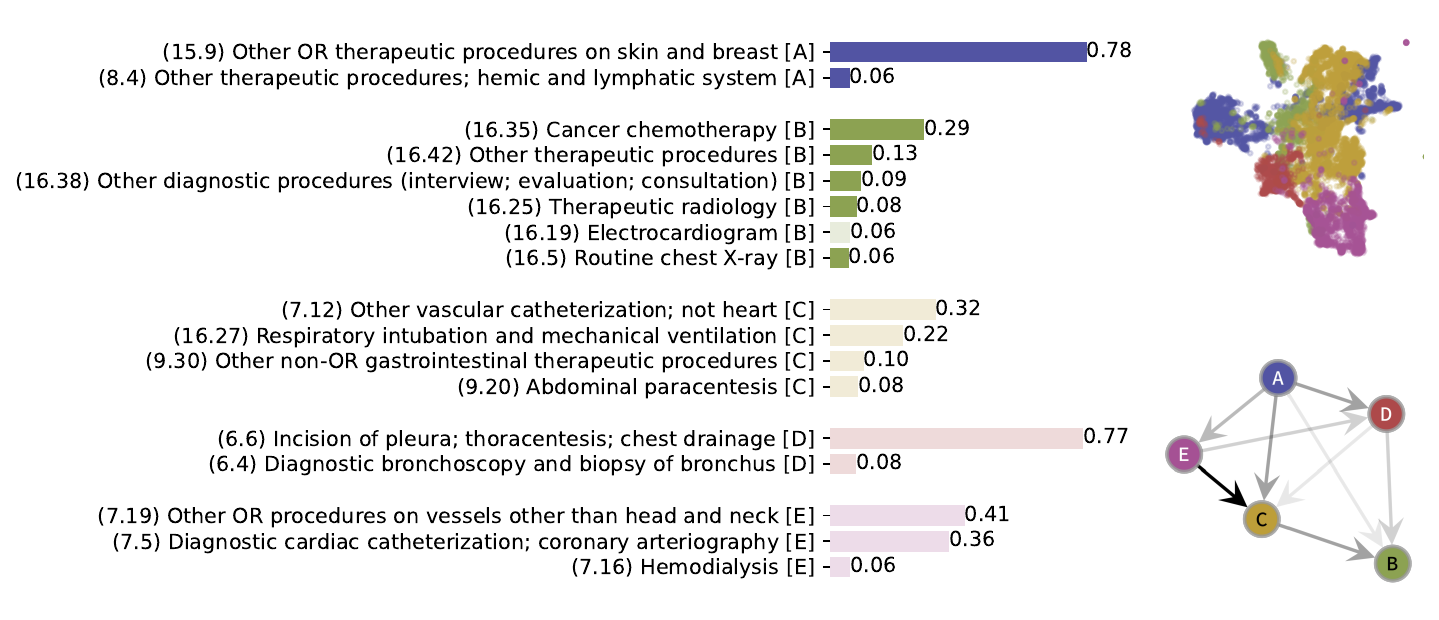}
        }
        \vspace{-0.3cm}
        \caption{\sffamily{\emph{Defrag} on the breast cancer dataset.}}
        \vspace{-0.3cm}
        \label{fig:1a}
        \center{
          \includegraphics[width=0.95\linewidth]{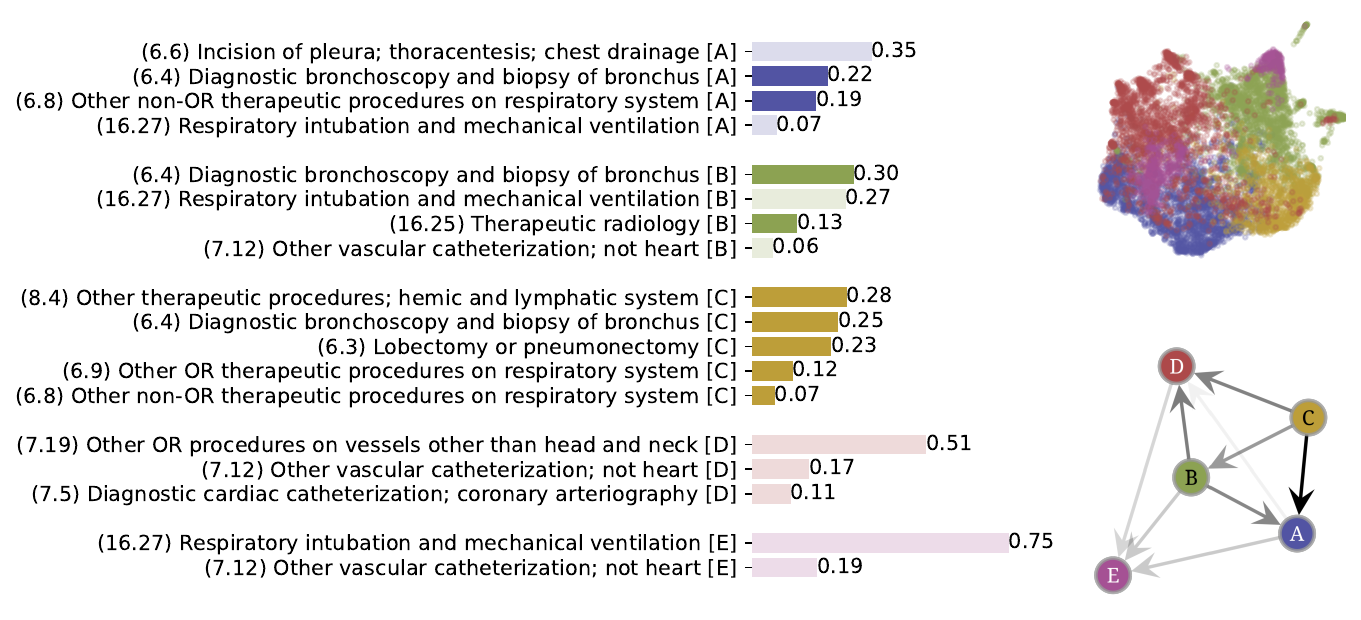}
        }
        \vspace{-0.3cm}
        \caption{\sffamily{\emph{Defrag} on the lung cancer dataset.}}
        \vspace{-0.3cm}
        \label{fig:1b}
        \center{
          \includegraphics[width=0.95\linewidth]{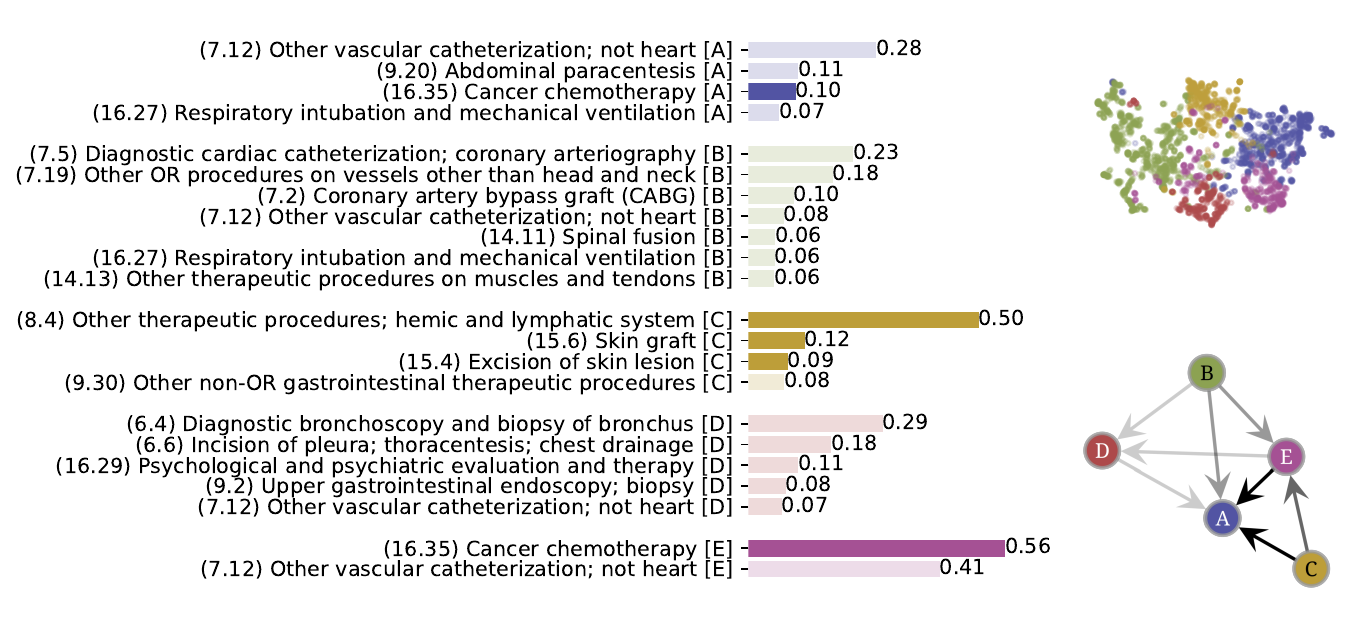}
        }
        \vspace{-0.3cm}
        \caption{\sffamily{\emph{Defrag} on the melanoma dataset.}}
        \label{fig:1c}
        \includegraphics[width=0.95\linewidth]{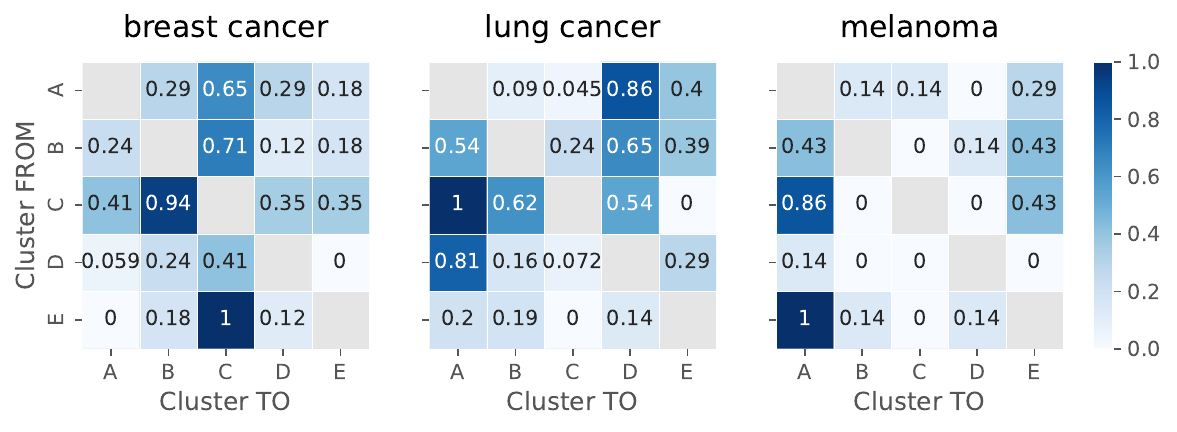}
        \caption{
            \sffamily{The raw adjacency matrices.}
        } 
        \label{fig:mimic_raw_adjacency_matrices}
    \end{subfigure}%
    \caption{
      Results from the MIMIC-IV experiments. Figures \ref{fig:1a}-\ref{fig:1c}: Left,
      TF-IDF-weighted procedure events (non-faded/most relevant, faded/least relevant); Top right:
      UMAP event embeddings; Lower right: Inferred pathway $G_{\to}$ (less frequent edges appear
      more faded). Figure \ref{fig:mimic_raw_adjacency_matrices}: The raw adjacency matrices of $G_{\leftrightarrow}$
      (normalised by the strongest edge weight) for each MIMIC-IV experiment. The y-axis indicates the
      source cluster, and the x-axis the destination cluster.
    } 
    \vspace{-0.3cm}
    \label{fig:mimic_summaries}
\end{figure}

Figure \ref{fig:mimic_summaries} depicts the results. We start by
analysing the unidirectional graph \(G_{\to}\) and the treatment event
distributions of each treatment cluster, as shown in Figures
\ref{fig:1a}-\ref{fig:1c}. For each cancer type, \emph{Defrag}
identifies modality-specific treatment procedure clusters. For example,
it detects cancer-related surgical procedures (clusters: ``A'' (breast),
``C'' (lung), and ``C'' (melanoma)) and adjuvant radio/chemotherapy
procedures (clusters: ``B'' (breast), ``B'' (lung), and ``A'' and ``E''
(melanoma)), which align with ESMO best-practice guidelines
\cite{Cardoso_2019,Vansteenkiste_2014,Michielin_2019}. It also uncovers
cancer-related diagnostic (non-treatment) procedures and unrelated
(non-cancer) clusters. We attribute the unrelated clusters to non-cancer
or ICU-related treatments and disease-agnostic procedures such as
catheterisation or mechanical ventilation.

Due to limitations of the MIMIC-IV dataset, we expect \emph{Defrag} to
find non-cancer related clusters, as a patient's cancer diagnosis might
not be their primary hospital admission reason. \emph{Defrag} reliably
identifies these clusters as: ``C, D, E'' (breast), ``D, E'' (lung),
``B, D'' (melanoma).

The summary in Figures \ref{fig:1a}-\ref{fig:1c} depict the simpler
unidirectional graph \(G_{\to}\) for simplicity; however, bidirectional
edges in \(G_{\leftrightarrow}\) are also important for observing the
relative proportions of adjuvant (after surgery) and neoadjuvant (before
surgery) therapies. Figure \ref{fig:mimic_raw_adjacency_matrices}
depicts raw adjacency matrices of \(G_{\leftrightarrow}\) to emphasise
these therapy proportions. In breast cancer, adjuvant (clusters ``A''
\(\to\) ``B'') and neoadjuvant (``B'' \(\to\) ``A'') chemotherapy and
radiotherapy are equally frequent. For lung cancer, adjuvant (``C''
\(\to\) ``B'') was more common than neoadjuvant (``B'' \(\to\) ``C'')
radiotherapy. Finally, for melanoma, adjuvant chemotherapy (clusters
``C'' \(\to\) ``E'') was frequently observed, while neoadjuvant
chemotherapy (``E'' \(\to\) ``C'') was not detected at all.

Based on these results, we conclude that Defrag can identify clinically
meaningful treatment pathways in AHRs. We also note that the inferred
pathways are not necessarily the same as the clinical pathways, as the
inferred pathways are based on the data, while the clinical pathways are
based on expert knowledge. Nonetheless, elements of the inferred
pathways investigated in this paper align with best-practice clinical
pathways in that they identified the same treatment modalities and their
relative temporal ordering (e.g., adjuvant and neoadjuvant chemotherapy
in breast and lung cancer, but adjuvant chemotherapy only for melanoma).
Furthermore, the inferred pathways also identified the relative
proportions of adjuvant and neoadjuvant therapies for each disease
within the population analysed, which can provide important insights
into the treatment patterns of cancer patients.

\hypertarget{testing-and-validation-experiments}{%
\subsection{Testing and Validation
Experiments}\label{testing-and-validation-experiments}}

This section tests \emph{Defrag} and other pathway inference techniques
on AHRs with known ground-truth pathways, as described in Section
\ref{testing-and-validation-framework}. We first run two demonstration
experiments to establish the task, then conduct numerous experiments to
assess the sensitivity of AHR-generation variables and evaluate
\emph{Defrag}'s performance against other methods.

\hypertarget{demonstration-experiments}{%
\subsubsection{Demonstration
Experiments}\label{demonstration-experiments}}

We run two experiments that differ only in pathway size: \(6\) or \(12\)
vertices. We use a single multinomial AHR variable with fixed support
\(|X| = 100\) and \(a = 3\). The Barabási--Albert algorithm generates
pathways with parameters \(m=1\), \(n=50\), \(p=0.1\), \(q=0\),
\(p_{i,adv} = 0.6\), and for \(1000\) patients. \emph{Defrag} is trained
with \(d=64\), \(16\) attention heads, \(4\) encoder and decoder layers
each, a feedforward dimension of \(64\), and dropout of \(0.2\). We
train the models for \(30k\) steps with a batch size of \(64\) and
optimised using AdamW \cite{Loshchilov_2019} with learning rate of
\(10^{-4}\). The encodings are clustered with HDBSCAN
\cite{McInnes_2017}. We also binarise the inferred pathway edges using a
threshold of \(0.2\). These parameters were determined through extensive
empirical experiments to balance model complexity and computational
efficiency.

\begin{figure}[t]
  \centering
  \includegraphics[width=0.8\linewidth]{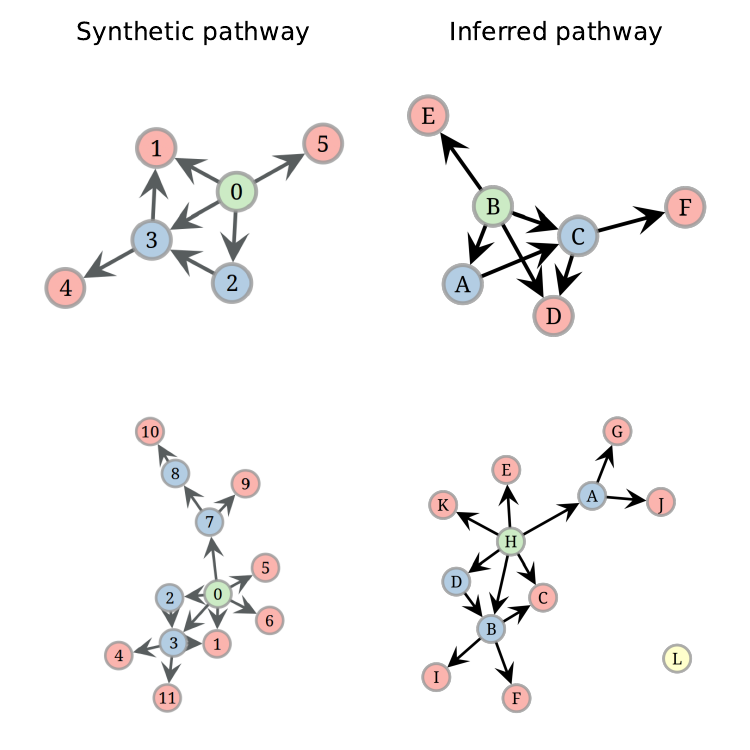}
  \caption{
    Results of two demonstration experiments with varying pathway sizes. Top row: $6$-vertex
    experiment; bottom row: $12$-vertex experiment. Left: synthetic, ground-truth pathway. Right: inferred
    pathway.
  }
  \vspace{-5mm}
  \label{fig:demo_experiments}
\end{figure}

Figure \ref{fig:demo_experiments} presents the results. \emph{Defrag}'s
inferred graph for the \(6\)-vertex experiment is isomorphic to the
ground-truth graph, while in the \(12\)-vertex experiment, one vertex is
disconnected, leading to incomplete inference. We quantify the
performance using the adjusted mutual information score (AMI)
\cite{Vinh_2009}, graph edit distance (GED) \cite{Sanfeliu_1983}, and
Weisfeiler-Lehman graph kernel (WLGK) \cite{Shervashidze_2011}. The AMI
results are \(0.79\) and \(0.82\), GED results are \(0\) and \(1\), and
WLGK of \(1\) and \(0.93\) for the \(6\)- and \(12\)-vertex experiments,
respectively. Note, it is critical to consider both AMI and GED/WLGK
scores when evaluating pathway inference methods, since AMI measures
semantic alignment of inferred treatments, while GED and WLGK measure
the structural alignment of the inferred and ground-truth pathways.
Also, in the remainder of the paper, we report GED-norm, which is the
GED score normalised by the number of nodes in the ground-truth graph,
enabling fair comparisons across varying graph sizes.

\hypertarget{benchmarking-and-evaluation}{%
\subsubsection{Benchmarking and
Evaluation}\label{benchmarking-and-evaluation}}

We vary four variables across \(192\) experiments to test
\emph{Defrag}'s performance in different circumstances: the number of
vertices in the graph (\({3, 5, 7, 9}\)) to adjust the pathway
complexity, the size of the fixed support (\(|X|: {100, 1000}\)) to vary
the vocabulary size, the exponent \(a\) of the \(\mathrm{Zipf}\)
variable (\({1.5, 2, 3, 4}\)) to control the noisiness of events, and
the number of variables (\({1, 2}\)) to vary the amount of information
available for each event observation. For each permutation of these
variables, we run three experiments with different seeds, and test
\emph{Defrag} against several baselines (HMM, LDA, PCA, Non-negative
matrix factorisation (NNMF), and Word2Vec (W2V)). For the methods that
generate vectors (PCA and W2V), we cluster the vectors with hierarchical
clustering to discretise the results. We also report a random baseline
that randomly assigns treatments to each event.

\begin{figure*}[t]
    \center{
      \includegraphics[width=\linewidth]{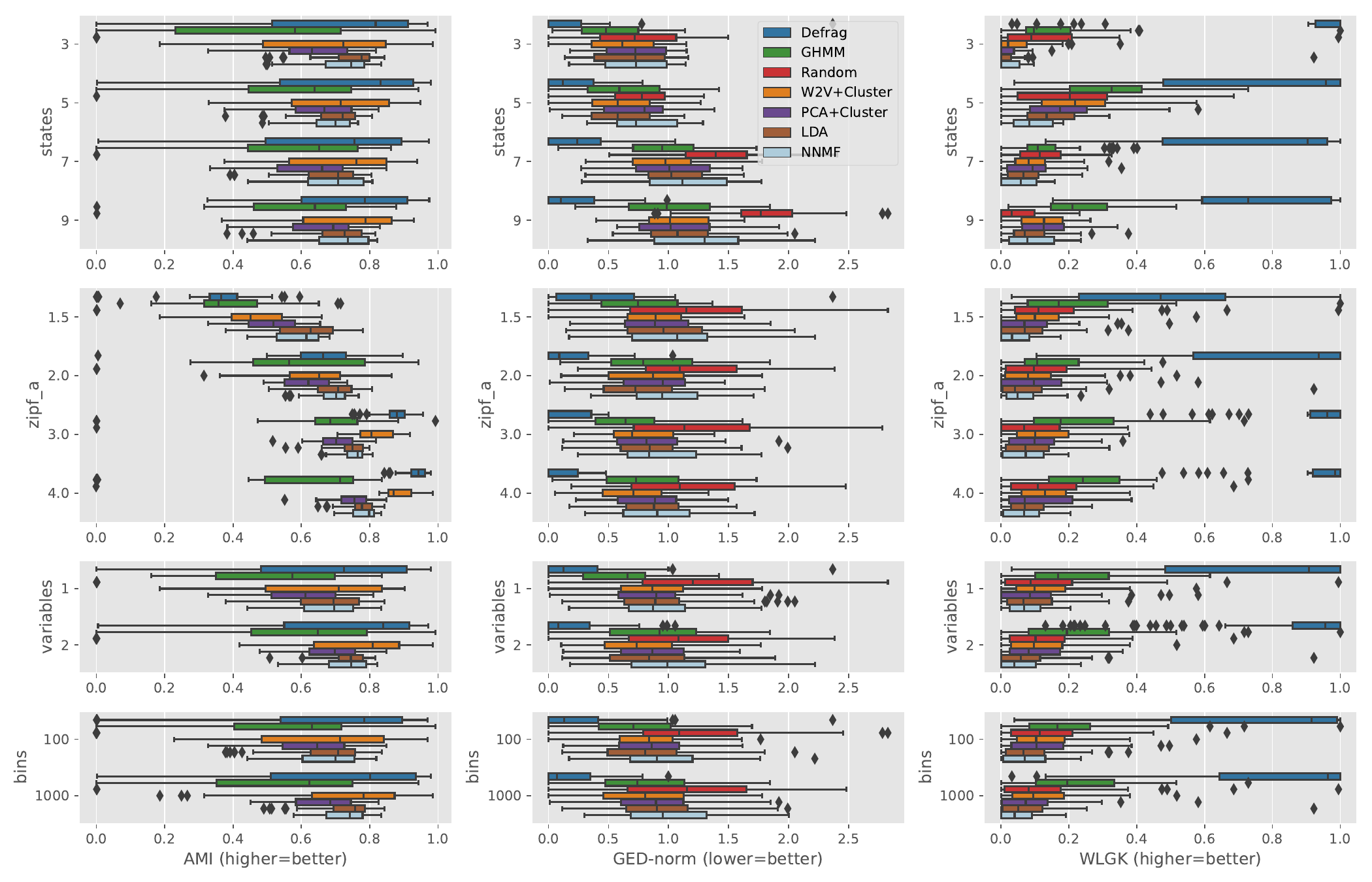}
    }
    \vspace{-5mm}
    \caption{
      Results of 192 pathway inference experiments comparing \emph{Defrag} to baselines.
      AMI (left), GED-norm (middle), and WLGK (right). The rows marginalise
      data synthesis variables to show how pathway inference is affected.
    }
    \label{fig:testing_defrag}
    \vspace{-5mm}
\end{figure*}

\begin{table}
    \caption{ Pathway inference performance of \emph{Defrag} and other baselines. The mean and
    standard deviation (in parentheses) of the metrics are reported. }
    \label{table:defrag_benchmarks}
    \center{
      \input{table_benchmarks.tex}
    }
\end{table}

Table \ref{table:defrag_benchmarks} demonstrates \emph{Defrag}`s
superior ability to infer the structure of the graph, as indicated by
the statistically significant\footnote{We use the Wilcoxon
signed-rank test \cite{Wilcoxon_1945} for statistical significance.}
improvements in GED-norm and WLGK scores. Figure
\ref{fig:testing_defrag} further explores synthetic data variables'
effects on pathway inference. Results suggest that \emph{Defrag}
performs better with less noisy AHR variables (higher \(a\)), smaller
graphs, more AHR variables, and simpler vocabularies (lower \(|X|\)).

\hypertarget{discussion-and-limitations}{%
\section{Discussion and Limitations}\label{discussion-and-limitations}}

The pathway inference method introduced in this paper can be used by
researchers and clinicians to undertake retrospective analysis of
treatment pathways in AHRs. This can be used to identify common pathways
and their temporal ordering, as well as to identify the relative
frequencies of different pathways and their components across different
patient cohorts. However, major gaps remain such as documenting and
quantifying the difference between the inferred pathways and the
clinical pathways that are based on expert knowledge, and cataloguing
the correspondence between observed AHR events and clinically-defined
treatment regimens. This is an important area for future research.

We did not explore using pre-trained features (e.g., Med2Vec
\cite{Choi_2016}) -- exploring the utility of such features for pathway
inference is an imporant area of future work. While we show that
\emph{Defrag} with STLO (Tables \ref{table:defrag_benchmarks} and
\ref{table:stlo-comparison}) is the most effective method for pathway
inference, exploring even more successful approaches is another
important area of future work, one which is now enabled thanks to the
synthetic data benchmarks and testing and validation framework
introduced in this paper.

We reiterate that the most suitable clustering method depends on the
dataset and the shape and density of encoded events. We find
hierarchical methods effective across our synthetic experiments (likely
due to the Barabási--Albert model) and MIMIC experiments, but
centroid-based methods may be better for other datasets. Furthermore,
since cluster optimisation and graph interpretability are not always
correlated, future work should explore other methods of utilising the
learned event vectors to enhance pathway inference.

We focus on population-level treatment pathways, but identifying rarer
pathways is also essential. While \emph{Defrag} can be used for this,
discerning weakly-weighted edges in \(G\) is challenging. Edge weights
indicate frequency, not significance; thus, rarer fragments may be
overlooked or undetected, as shown in Figure \ref{fig:demo_experiments}.
Future work should explore methods for identifying rare but important
edges in \(G\).

While this study focuses on addressing the major gaps in pathway
inference research (sequence modelling and evaluation), a comparitive
analysis of the network inference method used in combination with
\emph{Defrag}'s NN is an important area for future research.

Lastly, care should be taken when interpreting the edge weights in
\(G_{\to}\). These weights signify the relative directional strength of
edges, whereas the edge weights in bidirectional \(G_{\leftrightarrow}\)
signify the relative frequency of edges. \(G_{\to}\) offers a high-level
summary, while \(G_{\leftrightarrow}\) facilitates detailed cluster
transition analysis. Thus, \(G_{\to}\) should be analysed alongside
\(G_{\leftrightarrow}\) for a comprehensive view of edge behaviour.

\hypertarget{conclusion}{%
\section{Conclusion}\label{conclusion}}

This study presents \emph{Defrag}, an end-to-end, neural network-based
method for inferring treatment pathways in administrative healthcare
records (AHR). We test \emph{Defrag} on several AHR datasets,
demonstrating its effectiveness in identifying pathway fragments for
various cancer treatments and reconstructing pathways using synthetic
data. We open-source our method and synthetic data benchmarks to advance
clinical pathway inference research.

Future research should focus on interpreting inferred graphs and closing
the semantic gap between computationally-inferred and clinically-derived
pathways. Future pathway inference research should iterate on and
compare to \emph{Defrag}, utilising established benchmarks in the
testing and validation framework. Additionally, examining anomalies in
inferred pathways can help to inform clinical pathway development.

%% file: table_benchmarks.tex
\begin{tabular}{lrrr}
\toprule
     Method &             AMI &        GED-norm &            WLGK \\
\midrule
     \Defrag & $0.70$ ($0.26$) & $0.14$ ($0.28$) & $0.80$ ($0.29$) \\
       GHMM & $0.56$ ($0.24$) & $0.75$ ($0.32$) & $0.20$ ($0.15$) \\
     W2V+Cluster & $0.70$ ($0.19$) & $0.78$ ($0.25$) & $0.12$ ($0.11$) \\
PCA+Cluster & $0.64$ ($0.12$) & $0.88$ ($0.26$) & $0.09$ ($0.10$) \\
        LDA & $0.70$ ($0.10$) & $0.86$ ($0.27$) & $0.08$ ($0.10$) \\
       NNMF & $0.70$ ($0.09$) & $0.97$ ($0.30$) & $0.06$ ($0.05$) \\
\midrule
Random & $0.00$ ($0.00$) & $1.16$ ($0.59$) & $0.12$ ($0.14$) \\
\bottomrule
\end{tabular}

%% file: appendix.tex
\hypertarget{windowed-attention}{%
\section{Windowed-Attention}\label{windowed-attention}}

\begin{figure}[h]
    \begin{subfigure}{\columnwidth}
        \center{
          \includegraphics[width=\linewidth]{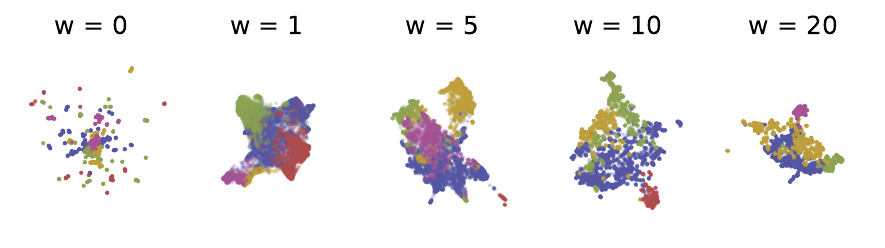}
        }
        \vspace{-0.3cm}
        \caption{\sffamily{UMAP embedding of $X_\mathbf{enc}$ on the breast cancer dataset for
        different values of $w$ in the windowed-attention.}}
        \label{fig:window_a}
        \center{
          \includegraphics[width=\linewidth]{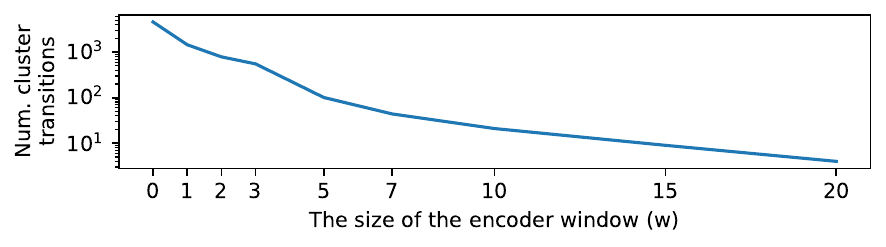}
        }
        \vspace{-0.3cm}
        \caption{\sffamily{The number of cluster transitions on the breast cancer dataset for
        different values of $w$ in the windowed-attention.}}
      \label{fig:window_b}
    \end{subfigure}
    \caption{
      This figure uses the breast cancer dataset to show how the behaviour of the learned
      semantic-temporal encoded event representations (top row) and the number of event-level
      cluster transitions in the inferred graph vary as the size $w$ of the encoder window varies. 
    }
    \label{fig:ablation_attention_window}
\end{figure}

The window size \(w\) of the attention mask in the Transformer encoder
controls the amount of temporal context for the learned
semantic-temporal encodings. Figure \ref{fig:ablation_attention_window}
shows how the value of \(w\) can result in three kinds of learned
embedding behaviour. Unlike Figure \ref{fig:mimic_summaries}, we omit
the TF-IDF-weighted cluster event distributions; these histograms relay
semantic meaning, which remains largely unchanged as \(w\) varies.
Instead, Figure \ref{fig:ablation_attention_window} focuses on showing
how the temporal meaning of clusters changes as \(w\) varies. This is
shown via the distribution of encoded events \(X_{\mathbf{enc}}\) in
Figure \ref{fig:window_a}, and the number of cluster transitions in each
inferred graph in Figure \ref{fig:window_b}, which is given as
\(\sum_{p \in V} \sum_{q \in V} w_{p, q, p \neq q}\) as per Equation
\ref{eq:defrag-infer}.

When \(w=0\) events can not attend to each other, and thus the learned
representations are semantic only (i.e., no temporal information is
used), with the representation space containing many regions of
high-density. The space must represent a total of \(|X|\) possible
discrete events.

When \(w = 1\), each event can attend to both of its neighbours from the
sequence, and thus temporal information is utilised which results in a
smoother representation space. If we assume that \(f_{\mathbf{enc}}\) is
not permutation invariant (which is a good approximation when \(w\) is
small), then the upper bound of unique neighbourhoods
(\(|X|!/(|X|-(2w+1))!\) or \(|X|!/(|X|-3)!\)) is significantly larger
than the number of events (\(|X|\)). This has the effect of increasing
the utilisation of the representation space, as indicated by the more
uniform UMAP embedding. However, since some adjacent events may appear
together quite often and thus have similar semantic-temporal
representations, the number of cluster transitions between adjacent
events decreases.

As \(w\) increases further, the neighbourhoods of adjacent events begin
to contain a greater proportion of redundant information. For example,
if \(w=5\), then the neighbourhoods of two adjacent events would contain
the same sequence of 10 events, a sequence which becomes increasingly
unlikely to belong to more than one patient. This redundancy is an
intentional consequence of the relative positional encoding
\cite{Shaw_2018} used in the Transformer's attention, which is required
to make \(f_{\mathbf{enc}}\) consider the sequence ordering. That is,
relative positional encoding makes \(f_{\mathbf{enc}}\)
translation-invariant. This would not be the case with absolute
positional encoding, (as in \cite{Vaswani_2017}) which would see the
number of event-neighbourhood permutations increase significantly and
make the representation space highly inefficient. Thus, as \(w\)
increases, the representation of individual events emphasises less of
the information about any one specific event and more of the information
about one specific patient. This all has the effect of making the
representation space less smooth, while continuing to reduce the number
of cluster transitions between adjacent events.

In practice, the choice of \(w\) will depend on the granularity of the
AHR dataset used. We found that inferred graphs were quite similar for
\(1 \leq w \leq 5\). As \(w\) becomes too large, there is not enough
information to infer a graph with sufficient detail for interpretation.

\hypertarget{relative-positional-encoding}{%
\section{Relative Positional
Encoding}\label{relative-positional-encoding}}

\begin{figure}[h]
    \includegraphics[width=\linewidth]{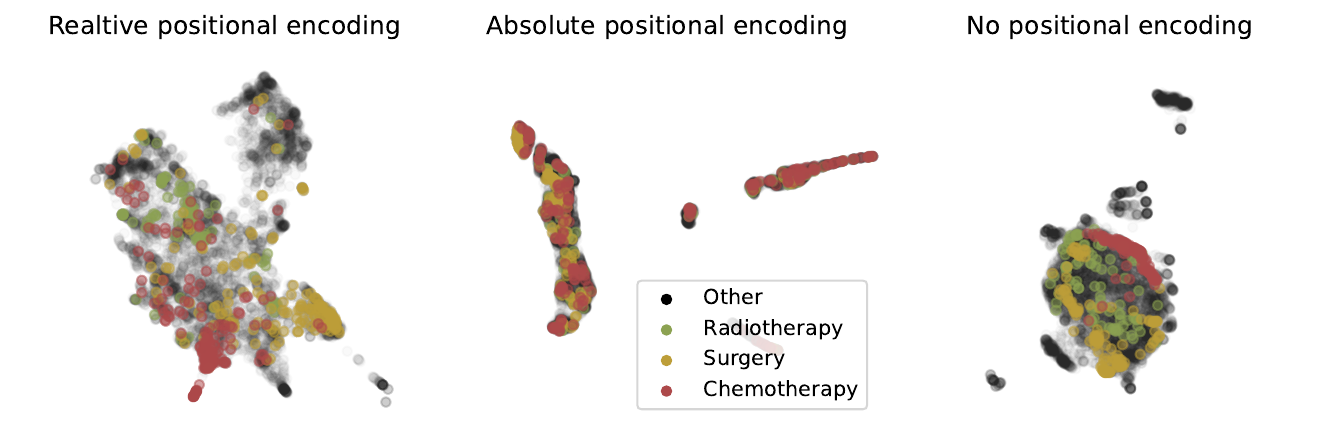}
    \caption{
      UMAP embeddings for the breast cancer experiment, with relative positional encoding (left),
      absolute positional encoding (middle), and no positional encoding at all (right). The colours
      depict events from three breast cancer treatment modalities. 
    } 
    \label{fig:positional_encoding_ablation}
\end{figure}

Figure \ref{fig:positional_encoding_ablation} shows UMAP embeddings of
the breast cancer experiment, each run with different positional
encoding schemes. While the experiment that used relative positional
encoding successfully separates three treatment modalities shown, the
absolute positional encoding and no positional encoding experiments do
not separate the modalities well at all. Separation of different
treatment modalities is critical for generating semantic-temporal
representations that are useful for pathway inference.

\hypertarget{stlo-discussion}{%
\section{STLO Discussion}\label{stlo-discussion}}

\hypertarget{comparison-with-other-self-supervised-objectives}{%
\subsection{Comparison with other self-supervised
objectives}\label{comparison-with-other-self-supervised-objectives}}

We provide pathway inference results for the STLO and other
self-supervised objectives in Table \ref{table:stlo-comparison}. Results
are reported on the 192 synthetic datasets from Section
\ref{benchmarking-and-evaluation}.

The results indicate that the STLO is the best self-supervised objective
for pathway inference. Notably, the next-best objective, SimCSE, is
competitive with the STLO in terms of pathway inference ability
(GED-norm and WLGK), but its ability to infer semantically accurate
clusters is significantly worse (AMI). This demonstrates that the STLO
has a unique ability to learn both semantically and temporally
meaningful representations of treatments.

\begin{table}
    \caption{Pathway inference performance of \emph{Defrag} when trained with various
    self-supervised learning objectives. The mean and standard deviation (in parentheses) of the
    metrics are reported.}
    \label{table:stlo-comparison}
    \center{
      \input{table_lossfn.tex}
    }
\end{table}

\begin{figure}[h]
    \begin{subfigure}{\columnwidth}
        \center{
          \includegraphics[width=\linewidth]{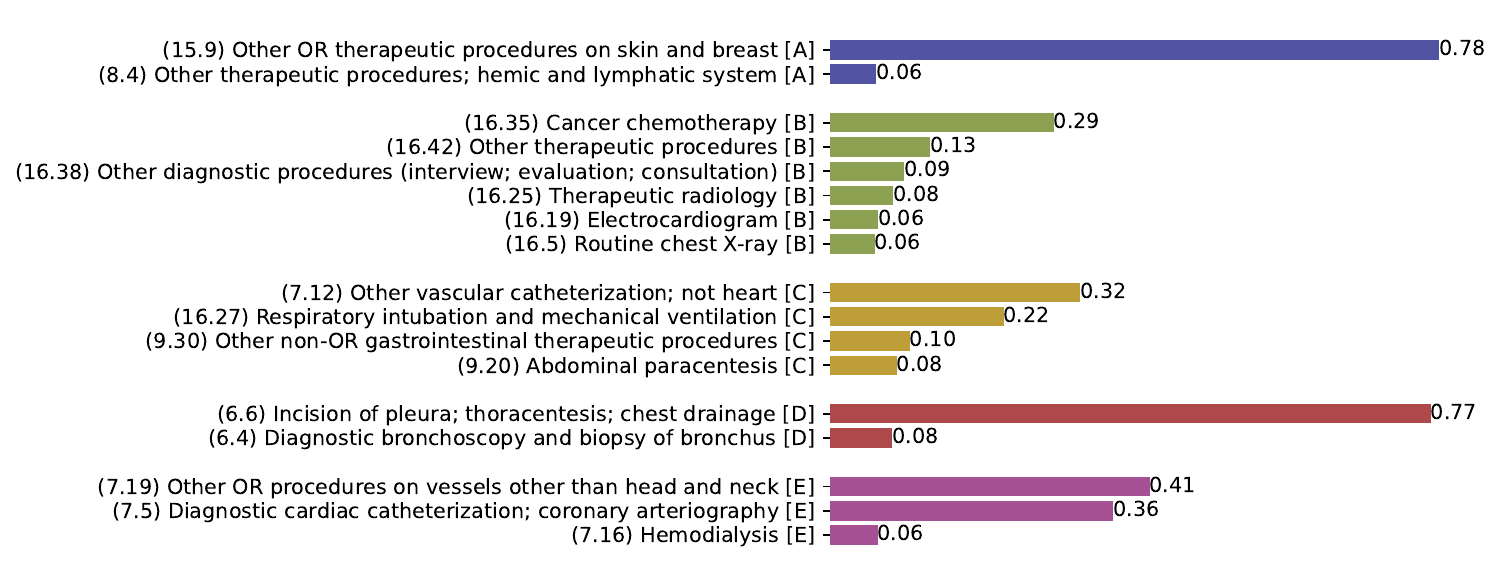}
        }
        \vspace{-0.5cm}
        \caption{\sffamily{Breast cancer experiment trained with STLO.}}
        \vspace{-0.2cm}
        \label{fig:stlo_cluster_stats}
        \center{
          \includegraphics[width=\linewidth]{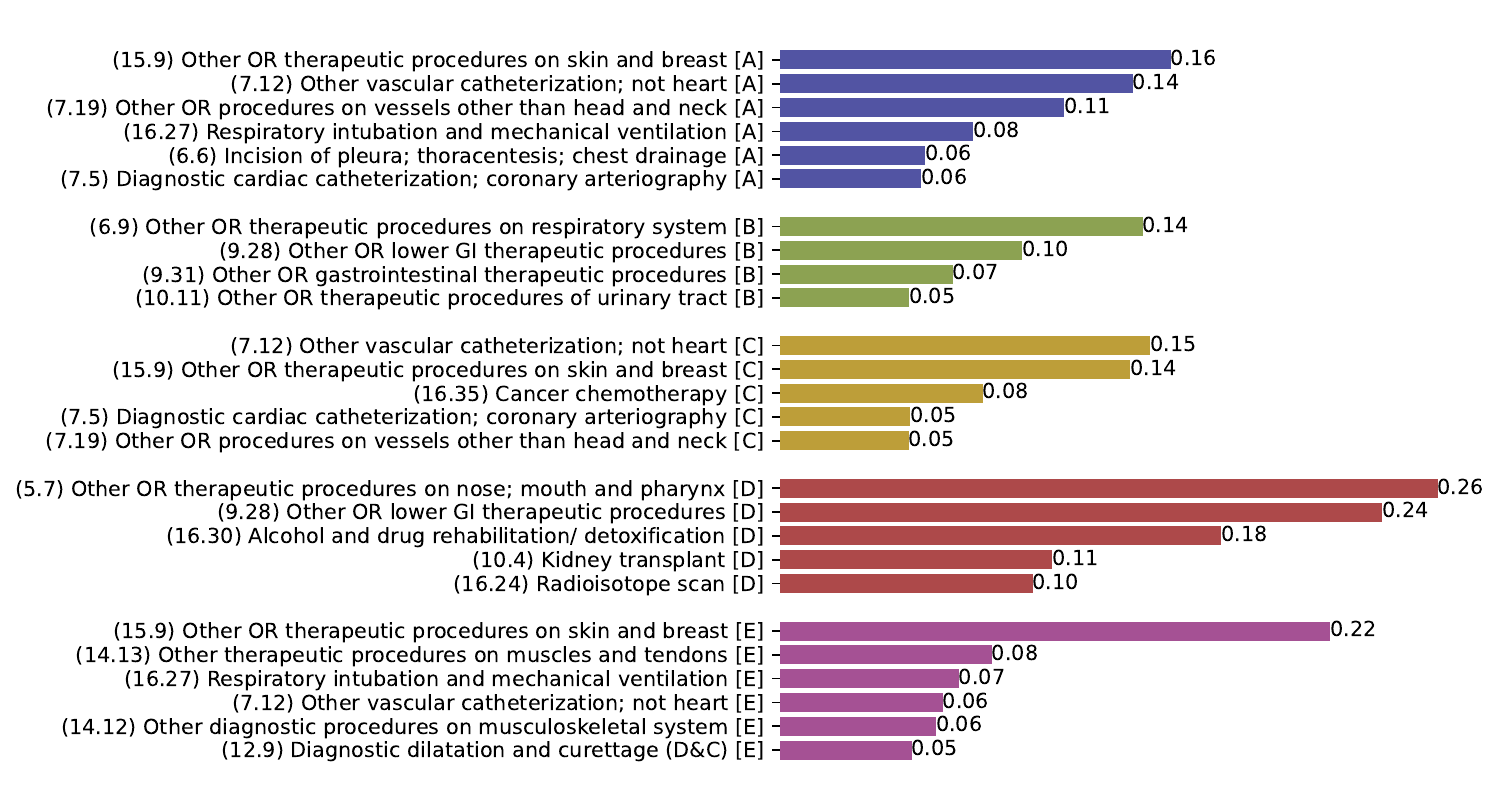}
        }
        \vspace{-0.5cm}
        \caption{\sffamily{Breast cancer experiment trained with an auto-encoding reconstruction
        objective on the decoder (mean squared error).}}
      \label{fig:mse_cluster_stats}
    \end{subfigure}
    \caption{
      TF-IDF-weighted distributions of treatment codes in the breast cancer experiment.
    }
    \vspace{-0.2cm}
    \label{fig:cluster_entropy}
\end{figure}

Furthermore, Figure \ref{fig:cluster_entropy} depicts the
TF-IDF-weighted cluster distributions of the breast cancer experiment,
each trained with a different loss function. The information entropy of
the STLO-based clusters is \input{figures/cluster_entropy_stlo.txt}
while the information entropy of the reconstruction-based clusters is
\input{figures/cluster_entropy_mse.txt}. Not only is the information
entropy of the STLO-based experiment significantly lower than the same
experiment trained with an auto-encoding reconstruction objective on the
decoder --- indicating that the clusters are not grouping temporally-
and semantically-similar treatments --- the histograms also show that
modality-specific treatments (e.g., ``Other OR therapeutic procedures on
skin and breast'') are isolated in the STLO-based experiment, but are
present in many different clusters in the reconstruction-based
experiment.

\hypertarget{two-treatment-assumption}{%
\subsection{Two-Treatment Assumption}\label{two-treatment-assumption}}

One inductive bias of the STLO is that it assumes that there are only
two treatments in the sequence. This is a pragmatic choice that is made
to simplify the training process since it is not possible to know how
many treatments there shall be since each patient is different and the
data contains no such treatment annotations. Furthermore, it is not
practical to determine how many treatments there are in the sequence in
an instantaneous, dynamic way during training. . While this choice is
not ideal, it is not a significant limitation of the STLO for the
specific purpose of learning semantic and temporal information about
treatments. Thus, we make the pragmatic choice to pick a static value
(two) and pursue several strategies to mitigate the consequences of this
decision, which we enumerate below.

\begin{enumerate}
\def\labelenumi{\arabic{enumi}.}
\tightlist
\item
  We train with short patient subsequences rather than entire patient
  sequences since shorter subsequences are more likely to contain a
  small number of treatments.
\item
  The STLO's two-treatment assumption is not a hard requirement. For
  example, should the short training subsequence contain three
  treatments instead of two, then the STLO will group two of the
  contiguous treatments together, and, to make the least bad decision
  about which treatments to merge, the model must use its knowledge of
  the semantic and temporal information of these treatments, which
  further contributes to the overall objective of learning semantic and
  temporal treatment information.
\item
  The `treatment' concept is abstract, not concrete. The STLO loss
  function is perhaps better conceptualised simply as a contrastive
  objective -- i.e., given a sequence of events, determine the point
  within the sequence that separates the events into two contiguous
  groups such that their within-group semantic and temporal similarity
  is maximised. Furthermore, this abstraction goes beyond the model,
  since the definition of a `treatment' varies across medical
  disciplines and data formats. For example, patients will likely
  receive fewer treatments in a emergency department visit, and more
  treatments across the course of a cancer treatment.
\item
  Finally, while the STLO loss at the decoder backpropagates through the
  encoder during training, it is not the only backpropagation path since
  the encoder and decoder paths diverge in the network architecture. We
  refer to \cite{Vaswani_2017} for a detailed description of the
  Transformer architecture. Therefore, the unique paths introduce the
  opportunity for the encoder and decoder to learn independent features
  that are relevant to the decoder's objective. Finally, we reiterate
  that the encoder's representations are intermediate in the Transformer
  and that intermediate features do not need to adhere to the training
  objective, rather they contribute useful information towards this
  goal.
\end{enumerate}

\hypertarget{encoder-decoder-setup}{%
\subsection{Encoder-decoder setup}\label{encoder-decoder-setup}}

\begin{figure}[h]
    \includegraphics[width=\linewidth]{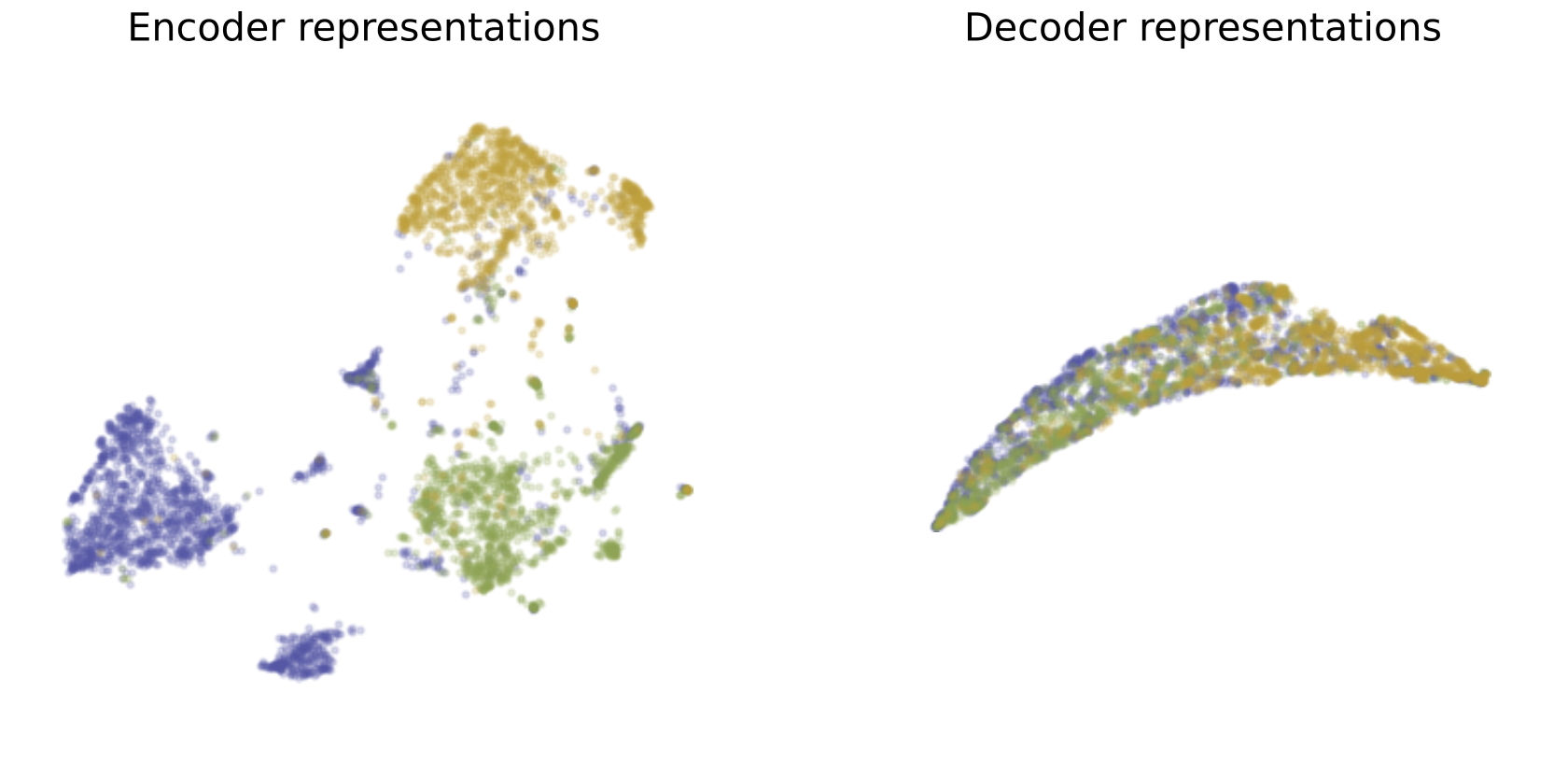}
    \caption{
       Empirical evidence from a synthetic data experiment containing three distinct treatments. The
      topology of the generated event representations suggests that clusters in the encoder's
      representations naturally align with ground-truth annotations. 
    } 
    \label{fig:enc-dec-rep}
\end{figure}

We choose the encoder-decoder setup of the Transformer (as opposed to a
decoder-only setup) since it allows us to decouple where STLO is applied
in the network from where the pathway inference is performed. This is
important since STLO is computed on the decoder, but the pathway
inference is conducted using the encoder's representations. Figure
\ref{fig:enc-dec-rep} shows that the encoder's naturally clusters the
data in alignment with the ground-truth annotations, which demonstrates
that the proposed neural network configuration is fit for purpose.

\hypertarget{cluster-optimisation}{%
\section{Cluster Optimisation}\label{cluster-optimisation}}

\begin{figure}[h]
    \includegraphics[width=\linewidth]{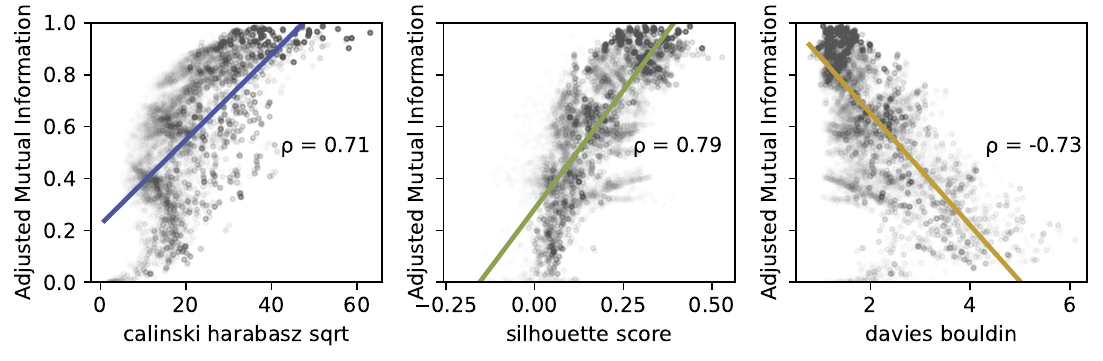}
    \caption{
      The correlation of unsupervised clustering metrics with AMI is examined. The square root of
      Calinski-Harabasz is plotted due to its quadratic scaling behaviour.
    } 
    \label{fig:cluster_optimisation}
\end{figure}

In the synthetic data experiments, optimal clustering on encoded events
is determined using a grid search, selecting the parameters for the
clustering method that optimises an unsupervised metric. No significant
practical differences were found among the following metrics:
Calinski-Harabasz \cite{Calinski_1974}, silhouette
\cite{Rousseeuw_1987}, and Davies-Bouldin scores \cite{Davies_1979}.
Figure \ref{fig:cluster_optimisation} depicts each of the grid-search
scores for all experiments in Section \ref{benchmarking-and-evaluation}.
All unsupervised scores roughly correlate with the supervised AMI
\cite{Vinh_2009}.

While automating the selection of clustering parameters was useful for
systematic experiments, this approach was not as useful for experiments
on MIMIC-IV data. This is because optimal clustering parameters depend
on the inductive biases of the metric and model, and are not necessarily
correlated with which partition over \(X_{\mathbf{enc}}\) yields the
most interpretable treatment clusters and inferred pathway \(G\). For
this reason, we chose to use identical clustering parameters across all
MIMIC-IV experiments to minimise the role of parameter selection in
influencing our analysis of the results.

In practice, an analysis should carefully examine several clustering
algorithms and clustering parameters (i.e., number and shape of
clusters) to inform a comprehensive analysis of treatment clusters. For
the MIMIC-IV experiments in Section \ref{mimic-iv-experiments}, we chose
to infer 5 clusters for the following reasons: 1) to be consistent
across each of the three experiments, 2) five clusters was the minimum
needed to infer sufficient granularity of the inferred pathway clusters,
3) inferred pathways containing more than five clusters become
increasingly difficult to interpret, and 4) When inferring more than
five clusters, we found that the encoded events were `over-clustered',
in that a single semantic cluster in the 5-cluster case was separated
into two functionally similar clusters in cases where the number of
clusters was greater than five. However, other values for the number of
clusters may be more suitable for other experiments -- to this end, we
reiterate that inferring more clusters yields more granular clusters,
vise versa.

\hypertarget{synthetic-ahr-plausibility}{%
\section{Synthetic AHR Plausibility}\label{synthetic-ahr-plausibility}}

\begin{figure}[h]
    \includegraphics[width=\linewidth]{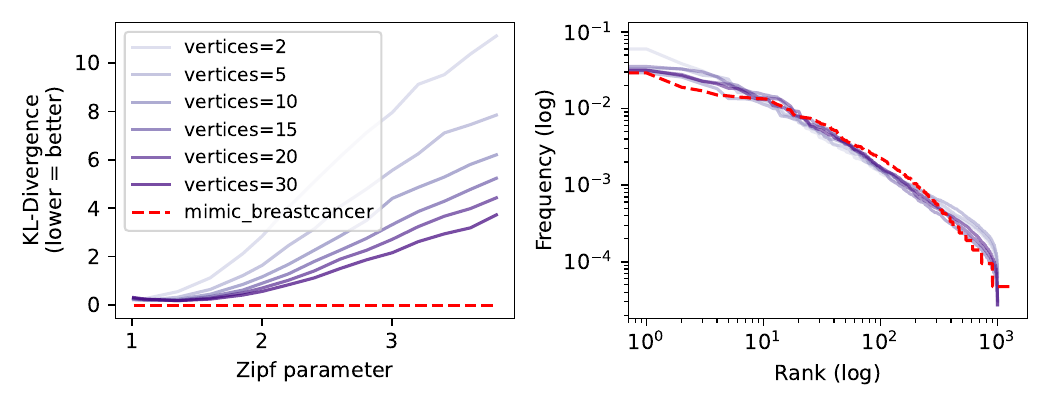}
    \caption{ 
      Left: KL-Divergence between synthetic data and MIMIC-IV distributions as the
      number of vertices in $G$ and Zipf parameter $a$ vary. Right: Rank-frequency distribution of
      the closest synthesised datasets for different-sized graphs and the MIMIC-IV
      distribution.  
    }
    \label{fig:kldivergence}
\end{figure}

Figure \ref{fig:kldivergence} compares the distribution of synthetically
generated AHR events with the ICD9 procedure codes in the MIMIC-IV
breast cancer dataset. Synthetic distributions are more aligned when
generated with more code variance (lower \(a\)) and larger graphs.
However, in general, the synthetic data can \emph{align} or
\emph{diverge} with real AHR datasets depending on the data generation
parameters. Alignment and divergence are both desirable properties of
such a testing and validation framework, since the goal is to simulate
different kinds of AHRs, not align too closely to any specific dataset.

%% file: table_lossfn.tex
\begin{tabular}{lrrr}
\toprule
               Method &             AMI &        GED-norm &            WLGK \\
\midrule
        \Defrag + STLO & $0.70$ ($0.26$) & $0.14$ ($0.28$) & $0.80$ ($0.29$) \\
      \Defrag + SimCSE & $0.59$ ($0.23$) & $0.18$ ($0.27$) & $0.74$ ($0.31$) \\
         \Defrag + MSE & $0.50$ ($0.27$) & $0.29$ ($0.36$) & $0.66$ ($0.35$) \\
\Defrag + Barlow & $0.48$ ($0.24$) & $0.36$ ($0.43$) & $0.57$ ($0.37$) \\
               \midrule
Random & $0.00$ ($0.00$) & $1.16$ ($0.59$) & $0.12$ ($0.14$) \\
\bottomrule
\end{tabular}

%% file: figures/cluster_entropy_stlo.txt
0.9, 2.0, 1.8, 0.9, 1.4 nats (mean: 1.4)\xspace

%% file: figures/cluster_entropy_mse.txt
2.1, 2.2, 2.1, 1.9, 2.1 nats (mean: 2.1)\xspace